\pdfoutput=1 

\documentclass[
     a4paper,
     ]{article}

\usepackage{IEK10} % defined in IEK10.sty
\usepackage{natbib}

\usepackage[export]{adjustbox}

\newcommand{\ronos}[1]{\text{RON\,$+$\,OS}}

\newcommand{\mytitle}{Graph Neural Networks for Temperature-Dependent Activity Coefficient Prediction of Solutes in Ionic Liquids}

\newcommand{\affil}{
  \begin{itemize}[leftmargin=3mm, itemsep=0mm]
        \item[$^a$]RWTH Aachen University, Process Systems Engineering (AVT.SVT), Aachen 52074, Germany %
		\item[$^b$]Delft University of Technology, Department of Chemical Engineering, Delft 2629 HZ, The Netherlands %
		\item[$^c$]Forschungszentrum J\"ulich GmbH, Institute for Energy and Climate Research IEK-10: Energy Systems Engineering, J\"ulich 52425, Germany%
		\item[$^d$]JARA-ENERGY, Aachen 52056, Germany
  \end{itemize}
}

\def\firstAuthor{Jan Rittig}
\newcommand{\myauthor}{
	Jan G. Rittig$^a$, 
	Karim Ben Hicham$^a$,
	Artur M. Schweidtmann$^b$, \\
	Manuel Dahmen$^{c}$, %
	Alexander Mitsos$^{d,a,c,*}$ %
}
\newcommand{\correspondingauthor}{
    Corresponding author, E-mail: amitsos@alum.mit.edu
}
\author{\myauthor}

\usepackage[hyphens]{url}
\usepackage[
  colorlinks,
  linkcolor=blue,
  citecolor=blue,
  urlcolor=blue,
  pdftitle={\mytitle},
  pdfauthor={\firstAuthor}
]{hyperref}
\usepackage[capitalise, nameinlink]{cleveref}
\crefname{table}{Tab.}{Tab.}

\begin{document}

	\thispagestyle{firststyle}
	
	\begin{center}
		\begin{large}
			\textbf{\mytitle}
		\end{large} \\
		\vspace{0.1cm}
		\myauthor
	\end{center}
	
	\vspace{-0.3cm}
	
	\begin{footnotesize}
		\affil
	\end{footnotesize}
	
	\vspace{-0.3cm}

	\section*{Abstract}
	
	Ionic liquids (ILs) are important solvents for sustainable processes and predicting activity coefficients (ACs) of solutes in ILs is needed.
	Recently, matrix completion methods (MCMs), transformers, and graph neural networks (GNNs) have shown high accuracy in predicting ACs of binary mixtures, superior to well-established models, e.g., COSMO-RS and UNIFAC.
	GNNs are particularly promising here as they learn a molecular graph-to-property relationship without pretraining, typically required for transformers, and are, unlike MCMs, applicable to molecules not included in training.
	For ILs, however, GNN applications are currently missing.
	Herein, we present a GNN to predict temperature-dependent infinite dilution ACs of solutes in ILs.
	We train the GNN on a database including more than 40,000 AC values and compare it to a state-of-the-art MCM.
	The GNN and MCM achieve similar high prediction performance, with the GNN additionally enabling high-quality predictions for ACs of solutions that contain ILs and solutes not considered during training.

	\vspace{0.1cm}

  %\newpage

%%%%%%%%%%%%%%%%%%%%%%%%%%%%%%%%%%%%%%%%%%%%%%%%%%%%%%%%%%%%%%%%%%%%%%%%%%%%%%%%%%%%%%%%%%
% Start of main Text

\section{Introduction}
Ionic liquids (ILs) have gained great interest in many chemical engineering applications due to their attractive physico-chemical properties such as negligible vapour pressure~\citep{Seddon.1997, Rogers.2003, Berthod.2018}.
A wide variety of ILs can be formed by combining anions, cations, and other structural groups, enabling tuning of ILs with respect to specific properties and resulting in a design space containing over one million potential binary ILs~\citep{Rogers.2003}.
Applications of ILs as solvents, catalysts, or electrolytes are vast, cf.~\citep{Seddon.1997, Rogers.2003, Plechkova.2008, Lei.2017} including separation processes~\citep{Brennecke.2001, Lei.2014}, biomass conversion~\citep{Zhang.2017}, and batteries~\citep{Galinski.2006}.
With these IL applications typically involving mixtures, the activity coefficient~(AC) is one of the most important properties as it accounts for intermolecular interactions and thus enables to model non-ideality of mixtures with ILs, cf.~\citep{Lei.2014, Han.2007, Zeng.2017, Chen.2021}.
For using ILs as solvents, particularly, the infinite dilution AC approximating non-ideal behaviour of solutions with high solvent and low solute concentrations is highly relevant.
To design ILs with application-specific properties, the AC therefore needs to be considered.
For exploring the large design space of potential ILs, computer-aided molecular design methods can be utilized, cf.~\citep{Karunanithi.2013, Peng.2017, Song.2018, Song.2019, Wang.2018}.
In relation to the large design space, however, the availability of experimental AC data is very limited.
Additionally, determining the AC experimentally for many IL candidates would be time-prohibitive and expensive.
Rather, models that enable fast and high-quality AC predictions are desired as they enable to explore a large number of IL candidates and are thus essential building blocks for computer-aided IL design.

Model-based AC prediction with thermodynamic equation-of-state methods and approaches like UNIFAC~\citep{Fredenslund.1975} and COSMO-RS~\citep{Klamt.1995} is well established in chemical engineering.
Such classical AC models have also been adapted for mixtures containing ILs.
For example, UNIFAC which maps the structural groups of molecules to ACs has been extended to ILs~\citep{Chen.2021, Song.2018, Lei.2009, Lei.2012, Roughton.2012, Dong.2020}, see, e.g., UNIFAC-IL~\citep{Song.2016}.
COSMO-RS, rooted in statistical thermodynamics, has also been applied and extended to the prediction of ACs of solutes in ILs~\citep{Chen.2021, Song.2016, Han.2018}.
For instance, Song et al. presented a modified COSMO-RS model calibrated to different molecular solute families by considering large amounts of experimental data which has been shown to produce improved results for IL solutions compared to standard COSMO-RS~\citep{Song.2016}.
However, classical AC prediction models also come with limitations.
UNIFAC can only be applied to molecules composed solely of those structural groups that have already been parameterized based on experimental data, and COSMO-RS has been found to have limited prediction accuracy for ACs of solutes in ILs~\citep{Chen.2021, Song.2016}.

Recently, methods from machine learning, namely matrix completion methods~(MCMs)~\citep{Jirasek.2020, Damay.2021}, transformers~\citep{Winter.2022}, and graph neural networks~(GNNs)~\citep{SanchezMedina.2022, Felton2022}, have been actively investigated as alternative approaches for predicting ACs, with MCMs also applied to IL solutions~\citep{Chen.2021}.
As the name suggests, MCMs encode solvents and solutes in a matrix, with AC of a specific solvent-solute combination as entries.
Since only few entries are filled by available experimental data, the matrix is typically sparse and has to be completed.
Recently, MCMs based on collaborative filtering~\citep{He.2017} have been utilized to AC prediction~\citep{Jirasek.2020, Damay.2021, Chen.2021}, relying on the concept that similarities between different rows and different columns, i.e., patterns, deduced solely from the given AC entries of a sparse solute-solvent matrix can be used to fill in the missing entries.
For example, Jirasek et al.~\citep{Jirasek.2020} utilized a MCM for predicting infinite dilution ACs at ambient temperature which performed favorably in comparison to UNIFAC.
Damay et al.~\citep{Damay.2021} extended this MCM model~\citep{Jirasek.2020} to also include a temperature-dependency of the infinite dilution AC by utilizing the Gibbs-Helmholtz relation.
For the prediction of temperature-dependent infinite dilution ACs of solutes in ILs, Chen et al.~\citep{Chen.2021} proposed a MCM that employs a neural recommender system, outperforming COMSO-RS~\citep{Song.2016} and UNIFAC-IL~\citep{Song.2016} on an extensive test set.
Despite the promising prediction accuracies of MCMs, their applicability is inherently limited to solvents and solutes for which at least one entry in the matrix is available.

Transformers, a ML method arising from the field of natural language processing, cf.~\citep{Vaswani.2017, Devlin.2018}, have also been utilized for molecular applications, e.g.,~\citep{Schwaller.2019, Rong.2020}.
Taking a sequence as input, for molecules typically consisting of SMILES strings~\citep{Weininger.1988}, transformers apply several feedforward neural network layers and attention mechanisms to learn relations within the input sequence relevant for mapping it to a property of interest, cf.~\citep{Rong.2020, Winter.2022}.
Very recently, Winter et al.~\citep{Winter.2022} proposed a transformer model to predict infinite dilution ACs of solutes in solvents at varying temperature based on SMILES strings, called SMILES-to-Properties-Transformer~(SPT).
Yet, transformers typically require large amounts of training data in the order of millions which are not available for many molecular properties.
Therefore, Winter et al. first generated about 10 million AC data points by means of COSMO-RS to pretrain the SPT and then used about 21,000 experimental AC values to fine-tune their model, thereby reaching high accuracy~\citep{Winter.2022}.
Generation of synthetic property data for this type of pretraining is, however, computationally expensive and limited by the availability and accuracy of existing models, which can thus hinder rapid development and extension of transformers for molecular property prediction. 

GNNs, another ML approach gaining great interest for molecular property prediction~\citep{Gilmer2017}, have only been applied to AC prediction very recently~\citep{SanchezMedina.2022, Felton2022}.
GNNs operate on graph-structured data; by representing molecules as graphs with atoms as nodes and bonds as edges, GNNs can be utilized to learn molecular properties directly from molecular graphs.
For AC prediction with GNNs, the solvents and solutes are thus represented as molecular graphs.
Since the molecular graph representation is generally applicable to molecules, GNNs can also process solutes and solvents that have not been included in the model training.
Thus, in contrast to MCMs, GNNs provide the possibility of enabling AC prediction for mixtures consisting of solutes and solvents not seen during model training.
Moreover, GNNs also enable data-scarce applications without pretraining, as required for transformers, see, e.g., our previous work~\citep{Schweidtmann.2020}.
Sanchez Medina et al.~\citep{SanchezMedina.2022} have recently utilized GNNs for the prediction of infinite dilution ACs at constant temperature for binary mixtures, achieving high prediction accuracies.
They also combined GNNs with classical AC prediction models such as COSMO-RS and UNIFAC into hybrid models, where a GNN learns to correct the error of a classical model~\citep{SanchezMedina.2022}.
Furthermore, Felton et al.~\citep{Felton2022} have presented DeepGamma, a deep learning model employing GNNs trained on millions of data points from COSMO-RS calculations, for predicting temperature-dependent ACs of binary mixtures at finite dilution.
Also, Qin et al~\citep{Qin.2022} have utilized COSMO-RS calculations for binary and additionally ternary solvent mixtures to develop and train SolvGNN, a GNN explicitly modeling intermolecular interactions for predicting ACs of multi-component solvent mixtures of different compositions at constant temperature. 
GNNs have thus already made promising advances in the field of AC prediction. 
However, the application of GNNs to IL solutions, the prediction of experimentally validated temperature-dependent infinite dilution ACs with GNNs, and the comparison to MCMs have not been investigated up to now.

We present a GNN for predicting temperature-dependent infinite dilution ACs of solutes in ILs\footnote{Code is openly available, see~\citep{GNN_GAMMA_IL_GIT}}.
In contrast to standard solvents, binary ILs constitute two disconnected but highly-attracted ionized molecules.
Therefore, we develop a GNN approach that takes three molecular graphs as input (the two IL molecules and the solute), and then learns a single continuous vector representation of the IL solution, referred to as molecular fingerprint.
The molecular fingerprint of the IL solution is then augmented with the temperature and mapped to the infinite dilution AC.
Thereby, the GNN architecture enables an end-to-end prediction from molecular graphs of IL solutions to ACs.
We thus extend current state-of-the-art GNN-based AC prediction models~\citep{SanchezMedina.2022, Felton2022, Qin.2022} to IL solutions and temperature-dependent infinite dilution ACs. 
We analyze the GNN prediction accuracy and compare it to state-of-the-art MCM methods~\citep{Chen.2021} for predicting the infinite dilution ACs of solutes in ILs.
In addition, we investigate the generalization capabilities of our GNN, i.e., we analyze if the GNN is able to predict experimentally validated ACs of solutions involving ILs and solutes that were not included in the training data set.

This work is structured as follows:
First, we describe the IL-solute data set (Section~\ref{sec:Dataset}) and present the GNN model for predicting the temperature-dependent AC of solutes in ILs (Section~\ref{sec:MaterialsMethods}).
We then present and discuss the prediction performance (Section~\ref{subsec:Results_Prediction}) and generalization capability (Section~\ref{subsec:Results_Generalization}).
Finally, we conclude our work and briefly discuss possible extensions of the presented GNN approach (Section~\ref{sec:Conclusion}).

\section{Data set}\label{sec:Dataset}
\noindent We use the data set of infinite dilution ACs at varying temperatures for IL solutions that was collected from the public ILThermo database~\citep{ILThermo} by Chen et al.~\citep{Chen.2021}.
The data set includes ILs and solutes with atoms of the types C, O, N, P, S, B, and halogens, whereas the charge of the ions within the ILs are $\pm$ 1 and each IL solution has at least ten data points, i.e., AC measurement at ten different temperatures, cf.~\citep{Chen.2021}.
The data set contains 215 ILs (consisting of 96 cations and 38 anions) and 112 solutes with a total number of 41,553 experimental $\gamma^\infty$ data points. 

\begin{table}
	\centering
	\caption{Data set of IL solutions from Chen et al.~\citep{Chen.2021} categorized by molecular classes of solutes. The number of data points per solute class are shown for the total data set, the training/validation, and the test set for two different model evaluation approaches (prediction and generalization).}
	\begin{tabular}{lrrrrr}
		\toprule
		\multicolumn{1}{l}{\multirow{2}[3]{*}{Solute family}}  &  \multicolumn{1}{c}{\multirow{2}[3]{*}{All}} & \multicolumn{2}{c}{Prediction} & \multicolumn{2}{c}{Generalization} \\
		\cmidrule(lr){3-4}\cmidrule(lr){5-6}
		\multicolumn{1}{c}{} & \multicolumn{1}{c}{} & \multicolumn{1}{c}{Train/Val} &  \multicolumn{1}{c}{Test} & \multicolumn{1}{c}{Train/Val} & \multicolumn{1}{c}{Test} \\
		\midrule
		Cl, F compounds &   1110 &   1017 &   93 &    698 &   412 \\
		acetic acid     &     29 &     24 &    5 &     29 &     0 \\
		acetonitrile    &    659 &    563 &   96 &    634 &    25 \\
		alcohols        &   5681 &   5051 &  630 &   5480 &   201 \\
		aldehydes       &    506 &    489 &   17 &    455 &    51 \\
		alkanes         &   6722 &   6065 &  657 &   5980 &   742 \\
		alkenes         &   3824 &   3383 &  441 &   3085 &   739 \\
		alkynes         &   2526 &   2275 &  251 &   2400 &   126 \\
		aromatics       &   6550 &   5996 &  554 &   4634 &  1916 \\
		cycloalkanes    &   3600 &   3169 &  431 &   3449 &   151 \\
		esters          &   1446 &   1284 &  162 &   1368 &    78 \\
		ethers          &   4236 &   3872 &  364 &   3667 &   569 \\
		ketones         &   2265 &   1979 &  286 &   2154 &   111 \\
		nitro alkanes  &    638 &    603 &   35 &    606 &    32 \\
		pyridine        &    429 &    398 &   31 &    404 &    25 \\
		terpenoids      &     46 &     46 &    0 &     46 &     0 \\
		thiophene       &    672 &    600 &   72 &    647 &    25 \\
		triethylamine   &     97 &     85 &   12 &     95 &     2 \\
		water           &    517 &    488 &   29 &    496 &    21 \\
		\midrule
		\textbf{Total}  &  \textbf{41553} &  \textbf{37387} & \textbf{4166} &  \textbf{36327} & \textbf{5226} \\  
		\bottomrule
	\end{tabular}
	\label{tab:Dataset}
\end{table}

We provide an overview of the data set categorized by solute classes in Table~\ref{tab:Dataset}.
The number of data points used for model training/validation and testing is shown with respect to two objectives: First, testing the prediction accuracy for IL solutions that contain molecules used for model training but in other combinations, and second, evaluating the prediction quality for generalization to IL solutions that contain molecules not used for model training.

For testing the prediction accuracy of our model, we use the same test set (about 10~\% of the total data set) as Chen et al.~\citep{Chen.2021} who ensured that IL-solute combinations that are included in the training/validation data set are not part of the test data set.
Note that we assume 10~\%, i.e., more than 4,000 data points, is sufficient to test the prediction quality of a model while retaining most of the valuable experimental data for training.
We apply a 90~\%/10~\% random split to the training/validation data set.

For evaluating the generalization capability to molecular structures not seen during training, we perform another split of the whole data set into training/validation and test set, with all IL-solute combinations in the test set including at least one molecule not included in the training/validation set.
Specifically, for the test set, we randomly select 5~\% from all unique SMILES~\citep{Weininger.1988} of both ILs and solutes (about 13~\% of all data points); the remaining data points are used as the training/validation set.
Analogously to the test set, we randomly select 5~\% of unique molecules from the training/validation set to create the validation set.
Each IL solution in the validation set and in the test set therefore contains at least one molecule not included in the training set and also not included in the test set or validation set, respectively.
Note that for each random training/validation split in the generalization analysis, the number of data points in the validation set typically varies because both the number of IL solutions a molecule is involved in and the corresponding number of temperature-dependent ACs may vary for different molecules.

\section{Methods \& Modeling}\label{sec:MaterialsMethods}

In this section, we first give a brief background on molecular graphs and GNNs, and then present our GNN model for AC prediction of solutes in ILs. 
We also provide insights on the MCM method we use for comparison, as well as on the training, implementation, and hyperparameter selection for both the GNN and the MCM model.

\subsection{Molecular graph}\label{subsec:MolecularGraph}

\noindent GNNs take molecular graphs as input.
Molecular graphs represent atoms of molecules with nodes/vertices and bonds between atoms with edges.
We denote nodes (vertices) with $v \in V$ and edges connecting two nodes $v,w \in V$ with $e_{vw}$.
In addition, each node and edge is assigned a feature vector that stores specific atom and bond information, respectively. 
The node feature vector is denoted by $\mathbf{f}^V(v)$ and contains, for example, information about the atom type or the formal charge of the atom.
Analogously, the edge feature vector is denoted by $\mathbf{f}^E(e_{vw})$ and typically includes information about the bond type.
The set of nodes and edges with the corresponding feature vectors describes the attributed molecular graph $G(m)= \{V, E, \mathbf{f}^V, \mathbf{f}^E\}$ for a molecule $m$.

We use the atom features shown in Table~\ref{tab:node_features} and the edge features illustrated in Table~\ref{tab:edge_features}; both are based on features reported in the literature~\citep{Gilmer2017} and our previous work on GNNs~\citep{Schweidtmann.2020}.
For the atom features, we additionally include the formal charge since ionic liquids are composed of ionized atoms or molecules.
Note that hydrogen atoms are not represented as nodes but are treated implicitly as atom feature by means of the count of hydrogen atoms bonded to a heavy atom.

\begin{table}[htb]
	\centering
	\caption{Atom features used for initializing node attributes. All features are implemented as one-hot-encoding.}
	\begin{tabular}{llc}
		\toprule 
		\multicolumn{1}{c}{ feature } & \multicolumn{1}{c}{ description } & dimension \\
		\midrule 
		atom type & type of atom (C, O, N, F, S, Cl, P, B, Br) & $9$ \\
		is in ring & whether the atom is part of a ring & $1$ \\
		is aromatic & whether the atom is part of an aromatic system & $1$ \\
		charge & formal charge of the atom (-1, 0, 1) & $3$ \\
		hybridization & sp, sp2, sp3, or sp3d2 & $4$ \\
		$\#$ Hs & number of bonded hydrogen atoms & $4$ \\
		\bottomrule
	\end{tabular}
	\label{tab:node_features}
\end{table}

\begin{table}[htb]
	\centering
	\caption{Bond features used for initializing edge attributes. All features are implemented as one-hot-encoding.}
	\begin{tabular}{llc}
		\toprule 
		\multicolumn{1}{c}{ feature } & \multicolumn{1}{c}{ description } & dimension \\
		\midrule 
		bond type & single, double, triple, or aromatic & $4$ \\
		conjugated & whether the bond is conjugated & $1$ \\
		is in ring & whether the bond is part of a ring & $1$ \\
		\bottomrule
	\end{tabular}
	\label{tab:edge_features}
\end{table}

\subsection{Graph neural networks}\label{subsec:GNNs}

\noindent GNNs~\citep{Gori2005, Scarselli2009} have recently been widely applied for molecular property prediction, e.g., in~\citep{SanchezMedina.2022, Gilmer2017, Schweidtmann.2020, Coley2017, Kearnes2016}.
GNNs learn to map the molecular graph to a property of interest in an end-to-end supervised training.
We show the structure of the GNN we use for predicting the AC at varying temperatures for ionic liquids in Figure~\ref{fig:model_architecture}.
The graph-to-property structure of GNNs is based on two phases: a message passing phase and a readout phase~\citep{Gilmer2017}.

In the \emph{message passing} phase, structure information within the molecular graph is encoded by means of graph convolutions. 
Graph convolutions are neural network layers that operate on the attributed molecular graph (cf. Section~\ref{subsec:MolecularGraph}).
Specifically, the feature vector of each node $v$ within the attributed molecular graph is considered individually and iteratively updated based on the feature vectors of the nodes and edges in the neighborhood, $N =\left\{w \mid e_{v w} \in E, w \neq v\right\}$. 
The updated feature vector of a node $v$ after passing a graph convolutional layer $l$ is typically referred to as hidden state $\mathbf{h}_v^l$.
Note that the atom features are utilized to initialize the hidden state, i.e., $\mathbf{h}_v^0 = \mathbf{f}^V(v)$.
The update process of a node hidden state within a graph convolution layer $l$ can be depicted as message information from the neighborhood passed to a node $v$ and is denoted by 
\begin{equation*}
	\mathbf{h}_v^{l} = \mathbf{U}_l(\mathbf{h}_v^{l-1}, \sum_{w \in N(v)} \mathbf{M}_{l}\left(\mathbf{h}_{w}^{l-1}, \mathbf{f}^{E}\left(e_{v w}\right)\right)),
\end{equation*}
where a message function $\mathbf{M}_l$ produces a message from the hidden state of a neighbor node $w$ of the previous graph convolutional layer, $\mathbf{h}_{w}^{l-1}$, and from the corresponding edge feature vector, $\mathbf{f}^E(e_{vw})$; then, the sum of all messages are passed to the node and are combined with the hidden state vector of the node from previous graph convolutional layer, $\mathbf{h}_{w}^{l-1}$, by applying an update function $\mathbf{U}_l$ which results in the updated hidden state of the node, $\mathbf{h}_v^l$.
For both the message and the update function multiple variations have been proposed, e.g., GCN~\citep{Hamilton2017}, GIN~\citep{Xu.2018}, higher-order methods~\citep{Morris2019, Flam-Shepherd2020}, and approaches including 3D information of molecules like atom distances and bond angles~\citep{Schutt2018, Unke.2019, Klicpera.06.03.2020, Zhang2020_MXM}.
By stacking multiple graph convolutional layers together, each node receives information from its neighbors, with each additional layer increasing the local neighborhood information passed to a node by one additional hop (from node to node) along an edge.
For example, in the second layer, the neighbors of a specific node have already received information from their respective neighbors in the previous layer which they can then pass on. 
The total number of graph convolutional layers $L$ employed thus describes the local information radius, the $L$-hop environment of an individual node encoded in the message passing phase. 

In the readout phase, the local structure information of the individual nodes is then aggregated into a continuous vector representation of the graph, the molecular fingerprint.
This aggregation of the single node hidden states is conducted by means of a pooling function, e.g., the sum, $\mathbf{h}_{\text{FP}} = \sum_{v \in V} \mathbf{h}_v^L$.
Taking the molecular fingerprint as an input, a feedforward neural network, typically a multilayer perceptron (MLP), is applied to predict a molecular property of interest, $\hat{p} = \text{MLP}(\mathbf{h}_{\text{FP}})$.

\subsection{Graph neural network for activity coefficient prediction}\label{subsec:GNN4AC}

\begin{figure}
	\centering
	\includegraphics[width=1\textwidth]{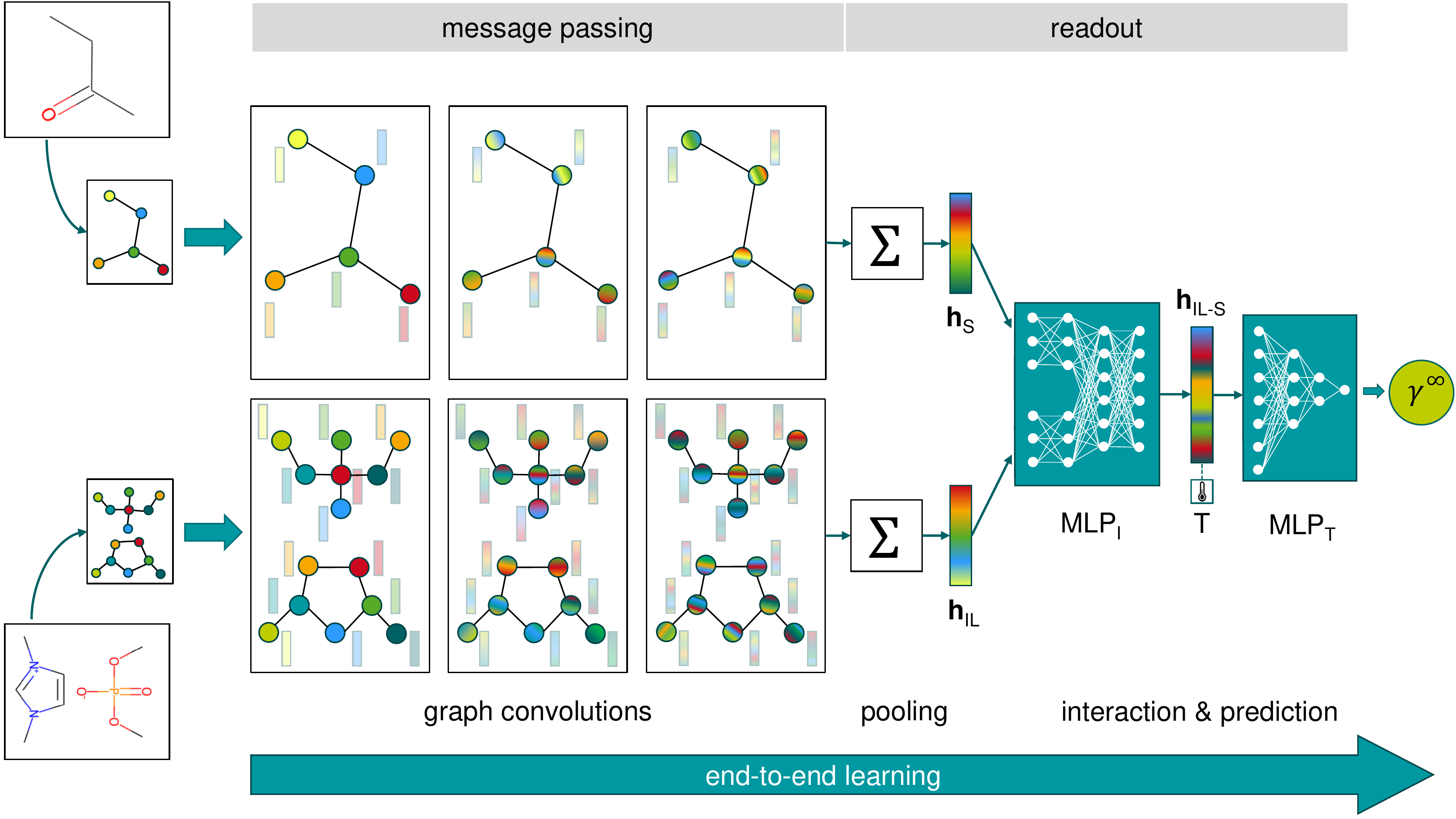}
	\caption{Graph neural network model for prediction of the temperature-dependent infinite dilution activity coefficient of solutes in ionic liquids.}
	\label{fig:model_architecture}
\end{figure}

\noindent Our GNN model for predicting infinite dilution ACs of solutes in ILs at varying temperatures is illustrated in Figure~\ref{fig:model_architecture}.
We first convert the molecules of an IL solution to attributed molecular graphs (cf. Section \ref{subsec:MolecularGraph}) that serve as input to the GNN.
In the message passing phase of the GNN, we use two separate graph convolutional layer channels, one for the molecular graph of the solute and the other one for the two molecular graphs, i.e., anion and cation, of the IL.
Thus, for the IL, the same graph convolutional layers are applied independently to the molecular graph of the anion and the molecular graph of the cation but the resulting hidden node state vectors are pooled into one IL fingerprint vector.
For the graph convolutions, we apply a gated recurrent unit (GRU) with the GINE-operator~\citep{Xu.2018, Hu.2019} that utilizes an $\text{MLP}_{\text{GINE}}$ to map the $\epsilon$-scaled hidden state of a node ($\epsilon$ being a learnable parameter) and the received information from the neighborhood (transformed by an activation function $\sigma$) to the updated hidden state, leading to the following update function:
\begin{equation*}
	\mathbf{h}_{v}^{l}=\operatorname{GRU}\left(\mathbf{h}_{v}^{l-1}, \sigma\left(\text{MLP}_{\text{GINE}} \left( (1 + \epsilon) \cdot
	\mathbf{h}^{l-1}_v + \sum_{w \in \mathcal{N}(v)} \mathrm{\sigma}
	( \mathbf{h}^{l-1}_w + \mathbf{f}_{e_{vw}})\right)\right)\right)
\end{equation*}
Note that here both the initial hidden node states and the edge features are linearly transformed by a learnable parameter matrix ($\theta$) to match the dimension of the following hidden states, i.e., $\mathbf{h}_v^0 = \theta_V \cdot \mathbf{f}^V(v)$ and $\mathbf{f}_{e_{vw}} = \theta_E \cdot \mathbf{f}^E(e_{vw})$, respectively.

After the GC layers, sum pooling is applied yielding the molecular fingerprint of the IL, $\mathbf{h}_{\text{FP,IL}}$, and the solute, $\mathbf{h}_{\text{FP,S}}$, respectively.
Then, the interactions between the IL and solute molecules are modeled with MLP$_{\text{I}}$, a MLP that transforms and concatenates the two molecular fingerprints.
The output of this interaction MLP is a combined IL-solute vector $\mathbf{h}_{\text{IL-S}}$, which is then concatenated with the min-max-normalized temperature $\text{T}_{norm}$ of the IL solution and subsequently fed into MLP$_{\text{T}}$, an MLP providing a prediction for the logarithmic value of the AC $\gamma^{\infty}$, i.e.,
\begin{equation*}\label{eq:gamma_MLP}
	\ln(\gamma^{\infty}) = \text{MLP}_{\text{T}}([\mathbf{h}_{\text{IL-S}}^T, T_{norm}]^T).
\end{equation*}

\subsection{Matrix completion method baseline}\label{subsec:MCM}

\noindent We re-implement an MCM model that has recently been presented for AC prediction of solutes in ILs by Chen et al.~\citep{Chen.2021}, as a state-of-the-art ML-based AC prediction benchmark for our GNN model.
Instead of molecular graphs, the MCM model takes one-hot encodings of ILs and solutes as input, i.e., each IL and solute is assigned an unique ID that is encoded in a one-hot-vector.
Additionally an unique ID for cation, anion, cationic family, and solute family (all four encoded as one-hot-vectors) are provided as input. 
For each of the six one-hot-encoded inputs, the MCM model employs several neural network layers, i.e., MLPs, to learn a continuous vector representation.
The six continuous vectors are then concatenated to form a single IL solution vector.
Analogously to the GNN model, the IL solution vector is then combined with the min-max-normalized temperature and mapped to $\ln(\gamma^{\infty})$ by another MLP.

\subsection{Ensemble learning}\label{subsec:EnsembleLearning}

\noindent For both the GNN and the MCM model we apply ensemble learning, a concept from machine learning that builds on the idea of averaging predictions of multiple models trained on different subsets of the training data set~\citep{breiman1996stacked, Breiman.1996, Dietterich.2000}.
Ensembles can increase the robustness of prediction models, e.g., by averaging out under- and over-predictions~\citep{Breiman.1996, Dietterich.2000}.

We apply ensemble learning by training multiple models on different splits of the IL-solute AC data set (cf. Section~\ref{sec:Dataset}).
Specifically, we split the data not used for testing randomly into a training and validation set before the training of each model.
After training, the outputs of all models are averaged to obtain the reported AC prediction.

\subsection{Implementation \& Hyperparameters}\label{subsec:Hyperparameters}

\noindent We implement our models in Python and utilize the geometric deep learning package PyTorch Geometric (PyG) developed by Fey \& Lenssen~\citep{Fey2019_PyGeo}. 
The annotated molecular graphs are generated with RDKit~\citep{rdkit}.
We provide the code and data used for training and testing open source, see~\citep{GNN_GAMMA_IL_GIT}.

The proposed GNN model exhibits several hyperparameters, which we tune in a two-step process.
In the first step, a grid search for the hyperparameters determining the GNN architecture is performed, varying the following hyperparameters within the respective ranges:
Graph convolutional type~$\in~\{\text{NNConv},~\text{GINEConv}\}$, number of graph convolutional layers~$\in~\{1,~2,~3\}$, usage of GRU in graph convolutions~$\in~\{\text{True}, \text{False}\}$, dimension of molecular fingerprint~$\in~\{64, 128\}$, number of layers in MLP-channels in interaction network~$\in~\{1,~2,~3\}$, activation function~$\in~\{\text{Leaky ReLU},~\text{ReLU}\}$.
Note that the number of neurons for the interaction MLP$_{\text{I}}$ is not varied and is set to 256 for all MLP-channel layers except for the last MLP-channel layer that has 128 neurons, followed by three interaction layers with dimension 256.
The structure of the MLP$_{\text{T}}$ is not varied and set to three layers with 257 (one additional dimension for the normalized temperature), 128, and 1 neurons.
The following training hyperparameters are applied: initial learning rate 0.001, learning rate decay of 0.8 with a patience of 3 epochs, batch size 64, maximum number of epochs 300, optimizer \textit{adam}, early stopping patience of 25 epochs, dropout rate in both MLPs of 0.05.
The first step of the hyperparameter search results in a final model architecture with the graph convolutional type GINEConv employed in two layers in combination with a GRU, a fingerprint dimension of 64, a number of layers in MLP-channels of 3, and Leaky RELU as activation function.
In the second step of the hyperparameter tuning, a grid search to fine-tune the GNN training parameters is conducted, i.e., varying the initial learning rate~$\in~\{0.01,~0.001,~0.0001\}$, the batch size~$\in~\{32,~64,~128~\}$, and the dropout rate~$\in~ \{0.1,~0.05,~0\}$.
We select the best model based on the validation error with a random split of the initial data set into training and validation sets.
This leads to an optimal initial learning rate of 0.001, a batch size of 64, and a dropout rate of 0.

Our MCM implementation uses the model structure with MLP blocks and the hyperparameter values proposed by Chen et al.~\citep{Chen.2021}. 
For training the MCM, we use a learning rate decay of 0.8 with a patience of 3 epochs and apply early stopping with a maximum of 300 epochs instead of a fixed number of 40 training epochs because we find the validation error to decrease in later epochs.

For choosing the ensemble size, we train 40 GNN and 40 MCM models and analyze the decrease in the validation MAE when step-wise combining more and more models of the same type starting from only 2 models.
We find that the validation MAE tends to stabilize between 30 to 40 models in case of both GNN and MCM and therefore consistently use ensembles of size 40 to generate the results. 

\section{Results \& Discussion}\label{sec:Results}

We first evaluate AC prediction for IL solutions, where both IL and solute were in the training set, however, in different combinations than in the test set (Section~\ref{subsec:Results_Prediction}).
Whereas the application of MCMs is inherently limited to this application scenario, GNNs can learn  molecular features from molecular graphs and thus can be applied to molecules not included in the training set. 
The latter application scenario is referred to as generalization, which we will investigate separately 
(Section~\ref{subsec:Results_Generalization}).

\subsection{Prediction of new IL-solute combinations}\label{subsec:Results_Prediction}

\noindent Table~\ref{tab:Prediction_newILsoluteSystems} shows the accuracies for predicting temperature-dependent ACs of new IL-solute combinations (both $\ln(\gamma^{\infty})$ and $\gamma^{\infty}$).
For example, the logarithmic form is relevant for calculating the chemical potential, whereas $\gamma^{\infty}$ is required when estimating vapor-liquid equilibrium.
Note that the logarithmic form is used for model training because it exhibits a more bell-shaped distribution and avoids variations in the order of magnitude as in the unscaled AC values.
The single model statistics present the accuracies on the training, the validation, and the test data averaged over 40 individual models of the same type.
For the ensemble of models, distinguishing training and validation accuracies is no longer possible (cf. Section~\ref{subsec:EnsembleLearning}), hence we provide the accuracies for training and validation data in one category.

\begin{table}[htbp]
	\centering
	\caption{Model prediction accuracies for new IL-solute combinations. Accuracy is provided by mean absolute error (MAE), root mean squared error (RMSE), coefficient of determination (R$^2$), and mean absolute percentage error (MAPE). For the single models, the standard deviation ($\pm$) across 40 different models is given.}
	\resizebox{\linewidth}{!}{%
		\begin{tabular}{lllllllll}
			\toprule
			\multicolumn{2}{c}{\multirow{2}[3]{*}{\shortstack[c]{Model setup}}} & \multicolumn{3}{c}{ln($\gamma^{\infty}$)} & \multicolumn{4}{c}{$\gamma^{\infty}$} \\
			\cmidrule(lr){3-5}\cmidrule(lr){6-9}
			\multicolumn{2}{c}{}   & \multicolumn{1}{c}{MAE} & \multicolumn{1}{c}{RMSE} & \multicolumn{1}{c}{R$^2$} & \multicolumn{1}{c}{MAE} & \multicolumn{1}{c}{RMSE} & \multicolumn{1}{c}{R$^2$} & \multicolumn{1}{c}{MAPE} \\
			\midrule
			\multicolumn{1}{c}{\multirow{5}[0]{*}{\shortstack[c]{GNN}}} 
			& single (train) & 0.027 $\pm$ 0.012 & 0.069 $\pm$ 0.015 & 1.00 $\pm$ 0.00 & 1.4 $\pm$ 0.7 & 16.7 $\pm$ 9.7 & 0.99 $\pm$ 0.02 & 2.74 $\pm$ 1.19\\
			& single (val) & 0.044 $\pm$ 0.009 & 0.109 $\pm$ 0.013 & 1.00 $\pm$ 0.00 & 2.0 $\pm$ 0.5 & 18.0 $\pm$ 6.9 & 0.99 $\pm$ 0.01 & 4.67 $\pm$ 1.03\\
			& single (test) & 0.093 $\pm$ 0.004 & 0.162 $\pm$ 0.006 & 0.99 $\pm$ 0.00 & 7.3 $\pm$ 1.4 & 67.1 $\pm$ 32.2 & 0.79 $\pm$ 0.28 & 9.59 $\pm$ 0.44 \\
			& ensemble (train/val) & 0.021 & 0.067 & 1.00 & 1.2 & 14.9 & 0.99 & 2.18 \\
			& ensemble (test) & 0.071 & 0.138 & 0.99 & 6.1 & 51.1 & 0.90 & 7.32 \\
			\midrule
			\multicolumn{1}{c}{\multirow{5}[0]{*}{\shortstack[c]{MCM}}} 
			& single (train) & 0.038 $\pm$ 0.002 & 0.087 $\pm$ 0.004 & 1.00 $\pm$ 0.00 & 2.9 $\pm$ 0.4 & 34.0 $\pm$ 4.9 & 0.96 $\pm$ 0.01 & 3.92 $\pm$ 0.23\\
			& single (val) & 0.050 $\pm$ 0.002 & 0.115 $\pm$ 0.009 & 1.00 $\pm$ 0.00 & 3.3 $\pm$ 0.7 & 32.5 $\pm$ 19.9 & 0.96 $\pm$ 0.04 & 5.30 $\pm$ 0.36\\
			& single (test) & 0.092 $\pm$ 0.002 & 0.157 $\pm$ 0.005 & 0.99 $\pm$ 0.00 & 5.8 $\pm$ 0.5 & 35.7 $\pm$ 4.2 & 0.95 $\pm$ 0.01 & 9.44 $\pm$ 0.29 \\
			& ensemble (train/val) & 0.030 & 0.084 & 1.00 & 2.72 & 33.5 & 0.96 & 3.20 \\
			& ensemble (test) & 0.076 & 0.138 & 0.99 & 5.0 & 30.7 & 0.96 & 7.70 \\
			\bottomrule
		\end{tabular}
	}
	\label{tab:Prediction_newILsoluteSystems}
\end{table}

The single GNN models achieve a mean absolute error (MAE) of 0.093 and R$^2$ of 0.99 on average for predicting $\ln(\gamma^{\infty})$ of the IL solutions in the test set.
For predicting $\gamma^{\infty}$, the MAE amounts to 7.3 and the mean absolute percentage error is 9.6~\%. 
In case of $\gamma^{\infty}$, the average R$^2$ value is visibly lower with a value of 0.79 which is caused by one of the GNN models failing to predict high $\gamma^{\infty}$ values, also causing the high standard deviation.
By using an ensemble of the single GNN models, i.e., averaging the predictions of all 40 models, the test set prediction accuracy further increases, yielding a reduced MAE of 0.071 for predicting $\ln(\gamma^{\infty})$.
Also, the MAE and mean absolute percentage error (MAPE) for predicting $\gamma^{\infty}$ are reduced to values of 6.1 and 7.3~\%, respectively.
Thus, we find an overall high prediction quality of the GNN, further enhanced by ensemble learning.

\begin{figure}
	\centering
	\includegraphics[width=0.8\textwidth, trim={0cm 0.5cm 0cm 0cm},clip]{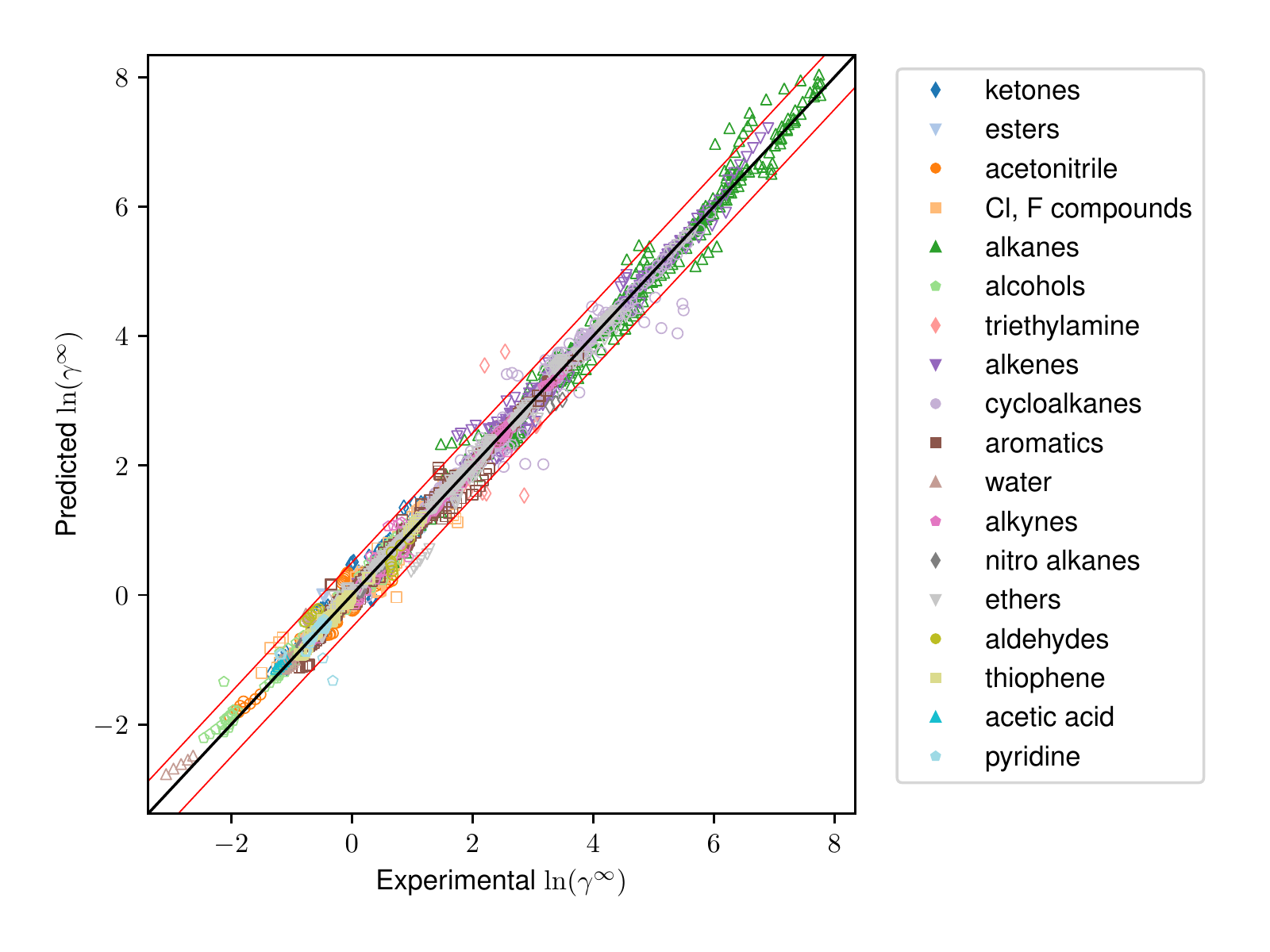}
	\caption{GNN model ensembling parity plot for test set with new IL-solute combinations. Red lines indicate $\pm$ 0.5 error range.}
	\label{fig:PredictionTest_ParityPlot}
\end{figure}

In Figure~\ref{fig:PredictionTest_ParityPlot}, we show the parity plot of experimental and predicted logarithmic ACs for the IL solutions of the test set when using the GNN ensemble.
A deviation of $\pm 0.5$ on the logarithmic scale is indicated by the red lines.
We find that the $\ln(\gamma^{\infty})$ predictions for almost all data points (i.e., 4109 out of 4166) are located within the $\pm 0.5$ error range.
The remaining data points have mostly slightly larger errors, with the highest absolute error being 1.357.
Furthermore, we do not find a systematic error.

\begin{figure}[htbp]
	\centering
	\includegraphics[width=0.8\textwidth]{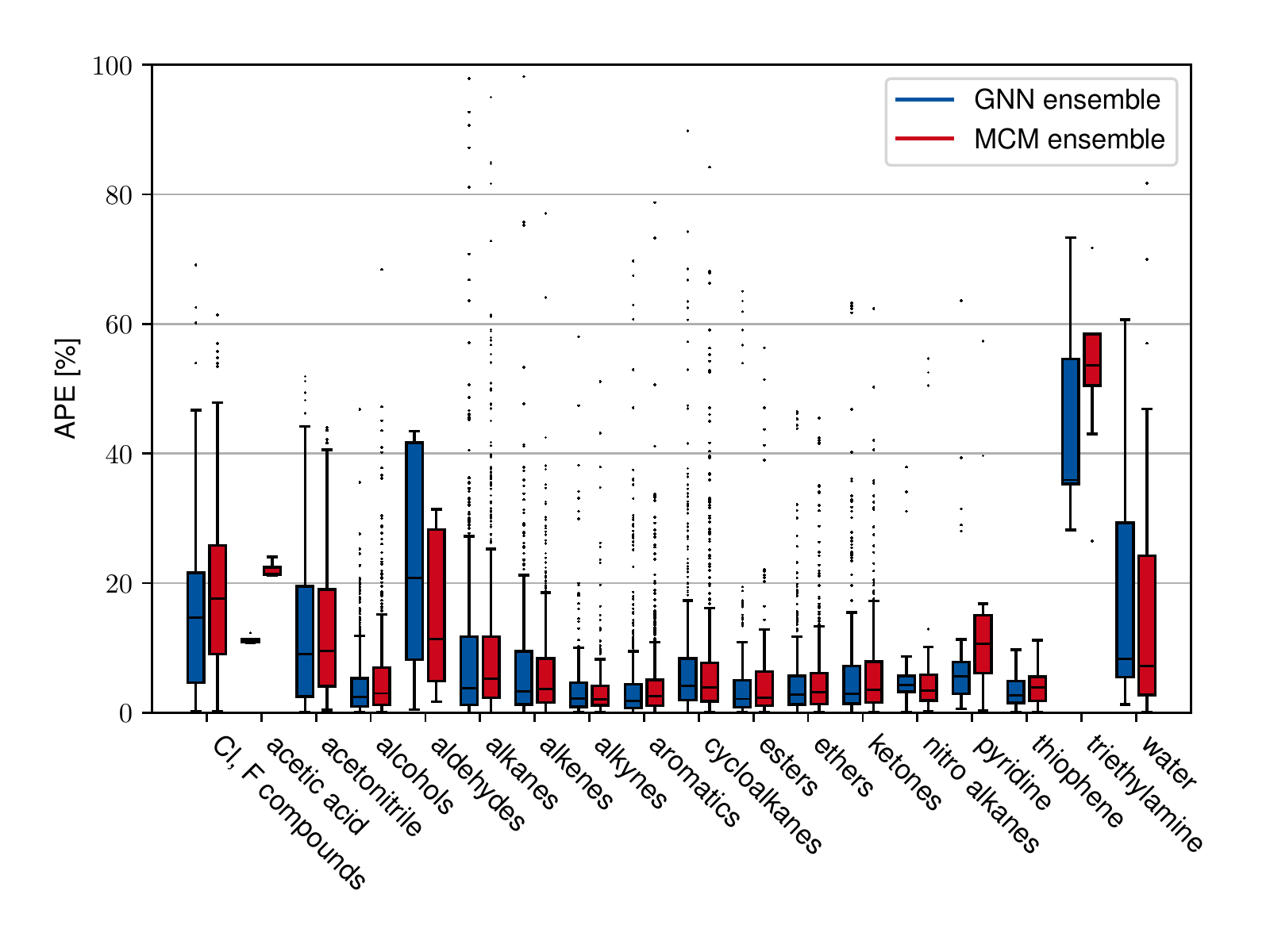}
	\caption{Absolute percentage error (APE) of GNN ensemble and MCM ensemble for predicting $\gamma^{\infty}$ of IL solutions in the test set categorized by solute families. For visualization, 13 outliers for the GNN ensemble and 12 outliers for the MCM ensemble with MAPE higher than 100~\% are not shown.}
	\label{fig:IL_Model_comparison}
\end{figure}

The comparison to the state-of-the-art MCM~\citep{Chen.2021} (cf. Section~\ref{subsec:MCM}) shows very similar performances on the same test set are achieved by both approaches, the ensemble of GNNs and the ensemble of MCM models (cf. Table~\ref{tab:Prediction_newILsoluteSystems}).
Note that the implemented MCM model has a validation root mean squared error (RMSE) of 0.115 averaged over 40 runs each with a randomly selected validation set which is similar to the 10-fold cross-validation RMSE of 0.136 reported by Chen et al.~\citep{Chen.2021} for their best model.
Also for the MCM, the application of ensemble learning considerably increases the prediction accuracy on the test set.

Chen et al.~\citep{Chen.2021} have shown that MCMs outperform classical AC prediction approaches such as the UNIFAC-IL model~\citep{Song.2016} and the calibrated COSMO-RS model~\citep{Song.2016}.
The GNN achieving competitive prediction accuracy to the MCM makes GNNs another promising alternative to UNIFAC-IL~\citep{Song.2016} and the calibrated COSMO-RS~\citep{Song.2016}.

Finally, we show the absolute percentage error (APE) of the GNN ensemble and the MCM ensemble for predicting $\gamma^{\infty}$ of IL solutions categorized by solute families in Figure~\ref{fig:IL_Model_comparison}.
For both the GNN ensemble and the MCM ensemble, the median APE lies below 20~\% for most solute families emphasizing the high prediction quality of both models.
A notable exception are solutions with triethylamine solutes with median APEs of 35.9~\% (MAPE of 77.3~\%) and 53.6~\% (MAPE of 71.7~\%) for the GNN ensemble and the MCM ensemble, respectively.
We explain the high error by the low number of data points for solutions containing triethylamines and the relatively low number of molecules with nitrogen atoms across all solute molecules in the data set compared to (oxygenated) hydrocarbons.
In addition, we observe outliers for many solute families (cf. Figure~\ref{fig:IL_Model_comparison}) which indicates that despite the low MAPEs both ML models generate strong mispredictions for a small fraction of IL solutions. 
More than 90~\% of all $\gamma^{\infty}$ values in the test, however, are predicted with an APE below 20~\% by both ensemble models.

\subsection{Generalization to new IL and solute molecules}\label{subsec:Results_Generalization}

\noindent We now present the results for predicting IL solutions that contain molecules not included in the training.
Table~\ref{tab:Prediction_newMolecules} shows the accuracies for predicting temperature-dependent ACs (both $\ln(\gamma^{\infty})$ and $\gamma^{\infty}$) categorized according to training, validation, and test data averaged over 40 single GNN models, as well as the ensemble learning results.
Again the logarithmic form is used for model training and for the ensemble model the training and validation error is aggregated into one category (cf. Section~\ref{subsec:EnsembleLearning}).
Comparing the prediction quality of the single GNN models to the ensemble, we again observe an accuracy increase for both training/validation and test set.
In the following, we thus focus on the ensemble results.

\begin{table}[htbp]
	\centering
	\caption{Model prediction accuracies for generalization to IL solutions containing at least one molecule not included in the training/validation set. Accuracy is provided by mean absolute error (MAE), root mean squared error (RMSE), coefficient of determination (R$^2$), and mean absolute percentage error (MAPE). For the single models, the standard deviation ($\pm$) across 40 different models is given.}
	\resizebox{\linewidth}{!}{%
		\begin{tabular}{lllllllll}
			\toprule
			\multicolumn{2}{c}{\multirow{2}[3]{*}{\shortstack[c]{Model setup}}} & \multicolumn{3}{c}{ln($\gamma^{\infty}$)} & \multicolumn{4}{c}{$\gamma^{\infty}$} \\
			\cmidrule(lr){3-5}\cmidrule(lr){6-9}
			\multicolumn{2}{c}{}   & \multicolumn{1}{c}{MAE} & \multicolumn{1}{c}{RMSE} & \multicolumn{1}{c}{R$^2$} & \multicolumn{1}{c}{MAE} & \multicolumn{1}{c}{RMSE} & \multicolumn{1}{c}{R$^2$} & \multicolumn{1}{c}{MAPE} \\
			\midrule
			\multicolumn{1}{c}{\multirow{5}[0]{*}{\shortstack[c]{GNN}}} 
			& single (train) & 0.103 $\pm$ 0.043 & 0.174 $\pm$ 0.060 & 0.99 $\pm$ 0.01 & 6.0 $\pm$ 2.2 & 64.5 $\pm$ 19.8 & 0.85 $\pm$ 0.08 & 10.7 $\pm$ 4.5  \\
			& single (validation) & 0.220 $\pm$ 0.053 & 0.357 $\pm$ 0.087 & 0.95 $\pm$ 0.03 & 9.6 $\pm$ 6.3 & 71.5 $\pm$ 58.2 & 0.74 $\pm$ 0.25 & 24.4 $\pm$ 6.1 \\
			& single (test) & 0.205 $\pm$ 0.030 & 0.306 $\pm$ 0.052 & 0.96 $\pm$ 0.01 & 6.4 $\pm$ 1.6 & 48.6 $\pm$ 26.1 & 0.82 $\pm$ 0.30 & 25.1 $\pm$ 5.4 \\
			& ensemble (train/validation) & 0.077 & 0.145 & 0.99 & 4.5 & 62.2 & 0.87 & 7.8 \\
			& ensemble (test) & 0.156 & 0.230 & 0.98 & 4.0 & 29.4 & 0.95 & 18.0 \\
			\bottomrule
		\end{tabular}
	}
	\label{tab:Prediction_newMolecules}
\end{table}

For predicting $\ln(\gamma^{\infty})$ of the IL solutions in the test set, the MAE of the ensemble amounts to 0.156 and the R$^2$ has a value of 0.98, indicating a high prediction quality.
The MAE value for $\gamma^{\infty}$ with 4.0 and the R$^2$ of 0.95 also correspond to high prediction accuracy.

Comparing the model for IL solutions with unseen molecules to the model for IL solutions with seen molecules (cf. Section \ref{subsec:Results_Prediction}), the MAE for $\ln(\gamma^{\infty})$ increases for both the single GNN models and the ensemble. 
That increase is not surprising since generalization to new molecules is considered inherently more difficult than predicting the AC for molecules already seen during training.
Since the respective test sets contain different data points, a direct quantitative comparison of the prediction accuracies, however, is not possible.
Overall, a high prediction quality is maintained for the generalization to unseen molecules.

\begin{figure}[htbp]
	\centering
	\includegraphics[width=0.8\textwidth, trim={0cm 0.5cm 0cm 0.5cm},clip]{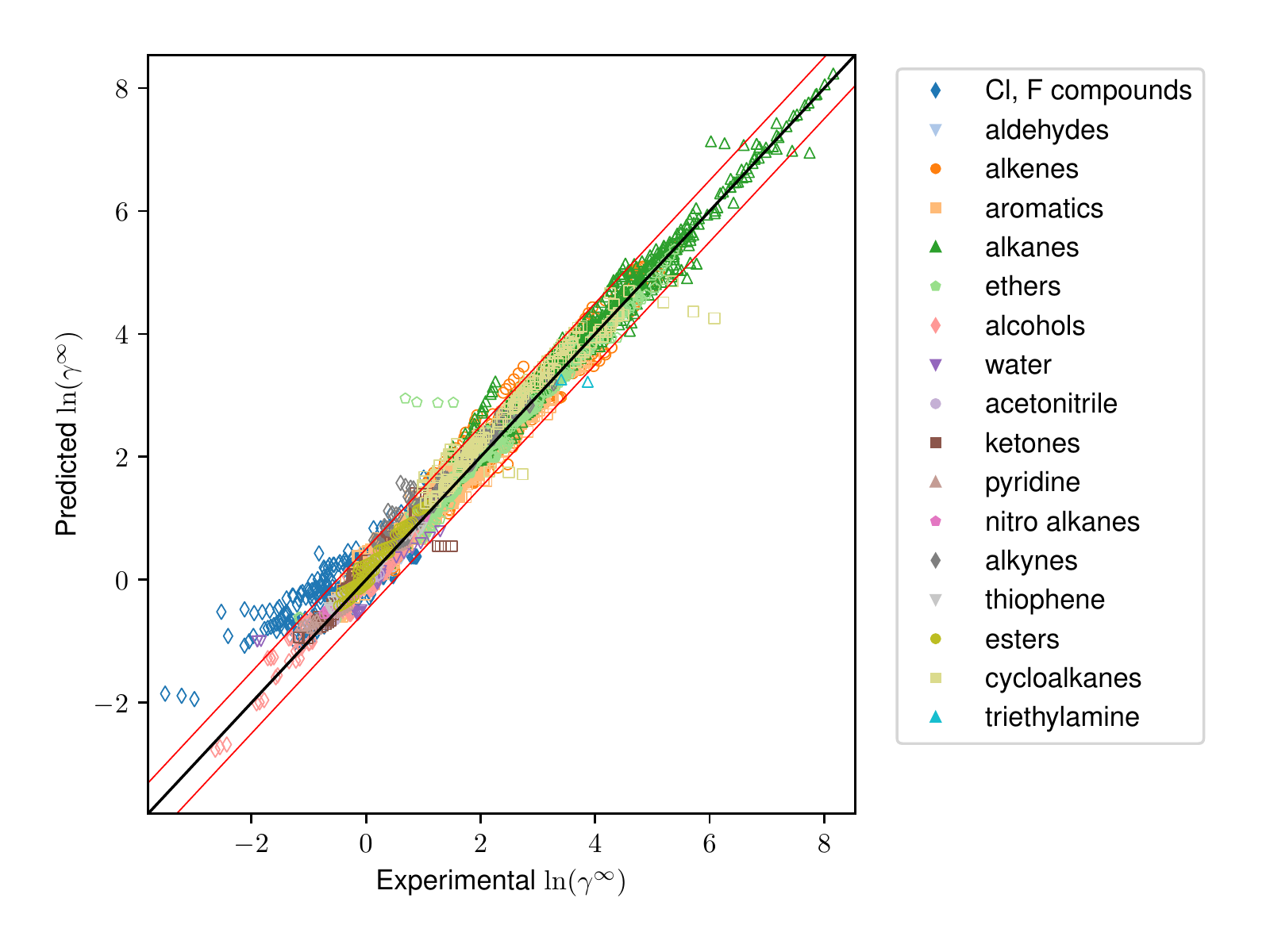}
	\caption{GNN model ensembling parity plot for generalization test set. Red lines indicate $\pm$ 0.5 error range.}
	\label{fig:GeneralizationTest_ParityPlot}
\end{figure}

We illustrate the experimental and predicted logarithmic ACs by the ensemble for the IL solutions within the test set for generalization by the parity plot in Figure~\ref{fig:GeneralizationTest_ParityPlot}.
A good match of the predicted values and the experimental values for $\ln(\gamma^{\infty})$ can be observed from the parity plot in Figure~\ref{fig:GeneralizationTest_ParityPlot}.
Most of the parity points lie within a deviation of $\pm 0.5$ on the logarithmic scale indicated by the red lines (Figure~\ref{fig:GeneralizationTest_ParityPlot}).
For the solute class of Cl, F compounds, the predictions tend to be too high for many data points.
Such deviation was not observed in the AC prediction of IL solutions with seen molecules (cf. Section~\ref{subsec:Results_Prediction}), hence we do not expect the error to be inherent to the GNN model architecture or noisy data.
We rather attribute the deviation to the fact that a large fraction of the Cl, F compound data points (37~\%) are located in the test set (cf. Table~\ref{tab:Dataset}) and thus fewer data points are available for training the model.
For the other solute classes, we observe a low number of outliers and do not find systematic prediction errors. 
Thus, the parity plot in Figure~\ref{fig:GeneralizationTest_ParityPlot} emphasizes the high prediction accuracy of our GNN in case of generalization to unseen molecules.

Overall, we find that GNNs provide high prediction quality and enable generalization for AC estimation of IL solutions with unseen molecules.
Predictions for molecules of classes with few data points available for training, however, should be taken with particular caution.

\section{Conclusion}\label{sec:Conclusion}

\noindent We present a GNN model for the prediction of temperature-dependent infinite dilution ACs of solutes in ILs. 
GNNs learn molecular properties based on a graph representation of molecules and have been successfully applied to AC prediction of solvents in solutes~\citep{SanchezMedina.2022, Felton2022, Qin.2022}.
We herein extend GNNs to AC prediction of IL solutions.
Specifically, we develop a GNN model that learns the infinite dilution AC as a direct function of IL and solute molecular graphs and the temperature.

The GNN model achieves high-accuracy AC predictions, superior to classical AC models such as COSMO-RS and UNIFAC-IL~\citep{Song.2016} and competitive with state-of-the-art MCMs for IL solutions~\citep{Chen.2021}.
Unlike MCMs, the GNN can also be applied to IL solutions with molecules not seen during model training, referred to as generalization.
We investigate the generalization capability by excluding some molecules from training and using them for testing.
Our results show that the GNN model allows for generalization with high accuracy, making it a highly promising constituent of computer-aided design of ILs. 

Future work could extend GNNs for AC prediction to IL solutions at finite dilutions, similar to~\citep{Felton2022}.
A further interesting direction is the combination of GNNs with classical AC models in form of hybrid models, cf.~\citep{SanchezMedina.2022, Jirasek.2021}.
Extending GNNs to provide chemically interpretable AC predictions with a quantified prediction uncertainty would also be highly desirable.

\section*{Acknowledgments}

\noindent This project was funded by the Deutsche Forschungsgemeinschaft (DFG, German Research Foundation) – 466417970 – within the Priority Programme ``SPP 2331: Machine Learning in Chemical Engineering''.
Simulations were performed with computing resources granted by RWTH Aachen University under project ``thes1105''.
AMS is supported by the TU Delft AI Labs Programme. 
MD received funding from the Helmholtz Association of German Research Centres.

\section*{Authors contributions}
\noindent \textbf{Jan G. Rittig}: Conceptualization, Methodology, Software, Formal analysis, Investigation, Writing - Original Draft, Visualization, Funding acquisition. 
\textbf{Karim Ben Hicham}: Methodology, Software, Formal analysis,  Investigation, Data Curation, Writing - Review \& Editing, Visualization.	
\textbf{Artur M. Schweidtmann}: Conceptualization, Writing - Review \& Editing, Funding acquisition.			
\textbf{Manuel Dahmen}: Conceptualization, Writing - Review \& Editing, Supervision.
\textbf{Alexander Mitsos}: Conceptualization, Writing - Review \& Editing, Supervision, Funding acquisition.

  %\clearpage
  %\newpage

  \bibliographystyle{apalike}
  \renewcommand{\refname}{Bibliography}
  \bibliography{literature.bib}

\end{document}